\documentclass[11pt]{article}

\usepackage[preprint]{acl}

\usepackage{times}
\usepackage{latexsym}

\usepackage[T1]{fontenc}

\usepackage[utf8]{inputenc}

\usepackage{microtype}

\usepackage{inconsolata}

\usepackage{graphicx}
\usepackage{longtable}
\usepackage{tabularx}

\usepackage[most]{tcolorbox}

\newtcolorbox{promptbox}[1]{
    enhanced,
    breakable,
    colback=gray!5,
    colframe=gray!45,
    coltitle=black,
    fonttitle=\bfseries,
    title={#1},
    boxrule=0.5pt,
    arc=2pt,
    left=8pt,
    right=8pt,
    top=6pt,
    bottom=6pt
}

\usepackage{color, colortbl}
\usepackage[table]{xcolor}
\usepackage{url}
\usepackage{subcaption}
\usepackage{textcomp}
\usepackage{soul}
\usepackage{multirow}
\usepackage{enumitem}
\usepackage{mathtools}
\usepackage{siunitx}
\usepackage{placeins}

\usepackage{booktabs} 
\usepackage{longtable,booktabs}

\usepackage{arydshln}
\definecolor{linkColor}{RGB}{6,125,233}
\definecolor{green}{rgb}{0.0, 0.65, 0.31}
\definecolor{bleudefrance}{rgb}{0.19, 0.55, 0.91}
\definecolor{ceruleanblue}{rgb}{0.16, 0.32, 0.75}
\definecolor{grey}{HTML}{969696}
\definecolor{violet}{HTML}{756bb1}
\definecolor{dgrey}{HTML}{01665e}
\definecolor{lgrey}{HTML}{5ab4ac}
\definecolor{dgreen}{HTML}{005a32}
\definecolor{purple}{HTML}{ae017e}

\definecolor{editCol}{HTML}{000000}
\definecolor{maskCol}{HTML}{c51b7d}
\definecolor{lrColor}{HTML}{8856a7}
\definecolor{trColor}{HTML}{d01c8b}
\definecolor{ctColor}{HTML}{4dac26}
\definecolor{brickred}{HTML}{f03b20}

\definecolor{sigone}{RGB}{222, 235, 247}    
\definecolor{sigtwo}{RGB}{158, 202, 225}    
\definecolor{sigthree}{RGB}{107, 174, 214}  

\definecolor{improveCol}{HTML}{4dac26}
\definecolor{worsenCol}{HTML}{d01c8b}

\definecolor{DarkBlue}{HTML}{00008B}
\definecolor{mscolor}{HTML}{01665e}
\definecolor{nmscolor}{HTML}{bf812d}
\definecolor{lgreen}{HTML}{ccece6}
\definecolor{dolive}{HTML}{308014}

\colorlet{tablerowcolor4}{gray!50} 

\newcommand*{\textlabel}[2]{%
  \edef\@currentlabel{#1}
  \phantomsection
  #1\label{#2}
}

\colorlet{tableheadcolor}{gray!25} 
\colorlet{tablerowcolor}{gray!10} 
\colorlet{tablerowcolor2}{gray!45} 
\colorlet{tablerowcolor3}{gray!12} 

\newcolumntype{a}{>{\columncolor{tablerowcolor}}r}
\definecolor{aicolor}{HTML}{018571}
\definecolor{occolor}{HTML}{ff7799}

\definecolor{aicolor}{HTML}{fc8d62}
\definecolor{occolor}{HTML}{253494}

\newif{\ifhidecomments}
  \hidecommentsfalse 
\ifhidecomments
    \newcommand{\olivia}[1]{}
    \newcommand{\agam}[1]{}
    \newcommand{\eshwar}[1]{}
    \newcommand{\koustuv}[1]{}
\else
    \newcommand{\olivia}[1]{\textbf{\small\sffamily{\textcolor{DarkBlue}{[#1 -- Olivia]}}}}
    \newcommand{\agam}[1]{\textbf{\small\sffamily{\textcolor{dgreen}{[#1 -- Agam]}}}}
    \newcommand{\eshwar}[1]{\textbf{\small\sffamily{\textcolor{orange}{[#1 -- Eshwar]}}}}
    \newcommand{\koustuv}[1]{\textbf{\small\sffamily{\textcolor{purple}{[#1 -- Koustuv]}}}}
  \fi

\colorlet{tableheadcolor}{gray!25} 

\definecolor{neutralCol}{HTML}{dd1c77}
\definecolor{neutralGreen}{HTML}{31a354}
\definecolor{NewBlue}{HTML}{1879ba}
\definecolor{bleudefrance}{rgb}{0.19, 0.55, 0.91}  
\definecolor{AfTrColor}{HTML}{0868ac}  
\definecolor{BfTrColor}{HTML}{a8ddb5}  

\definecolor{AfCtColor}{HTML}{b10026}  
\definecolor{BfCtColor}{HTML}{fd8d3c}

\graphicspath{ {figures/} }

\newcommand{\para}[1]{\vspace{0.2em}\noindent\textbf{#1}~}

\title{Secondhand Stress: How Stressful Disclosures Affect Psychosocial Wellbeing in Online Graduate Admissions Communities}

\author{
 \textbf{Veda Duddu},~
 \textbf{Agam Goyal},~
 \textbf{Sabelle Huang},~
 \textbf{Ayush Umashankar Prasad},~
 \textbf{Mingquan Yi},~
\\
 \textbf{Koustuv Saha}
\\
 Siebel School of Computing and Data Science\\  University of Illinois Urbana-Champaign, Urbana, IL, USA
\\
\texttt{\{vduddu2, agamg2, sabelle2, aup2, myi22, ksaha2\}@illinois.edu}
}

\begin{document}
\maketitle
\begin{abstract}
Online communities provide spaces for individuals navigating stressful life transitions to disclose difficulties and seek peer support. Yet encountering others' distress may also have psychosocial implications for community members. Drawing on stress crossover theory, we examine how engagement with peer distress during graduate admissions is associated with subsequent psychosocial expression and participation, and whether similar associations extend through downstream interactions. We analyze 171,981 posts and 1,130,413 comments from two Reddit communities across three admissions cycles, using propensity-score stratification and baseline-adjusted comparisons. Users who engaged with distress disclosures during application preparation subsequently exhibited higher classifier-estimated anxiety, depression, and stress expression, alongside greater commenting activity and lexical redundancy. These associations recurred across admissions cycles, while no significant psychosocial associations were detected among downstream respondents interacting with previously exposed users. Our findings highlight how greater distress-related expression can coexist with continued community participation, while distinguishing direct engagement with distress disclosures from subsequent social contact. We discuss implications for designing online communities that preserve opportunities for self-disclosure and peer support while remaining attentive to the experiences of those encountering distress-related content.
\end{abstract}

\section{Introduction}
Graduate admissions represent a consequential life transition, as applicants navigate uncertainty surrounding their educational and professional futures. Life transitions involve changes in roles, responsibilities, and expectations~\cite{schlossberg1981model}, while educational, employment, and relocation transitions can disrupt familiar sources of support~\cite{wendlandt2008addressing,mikal2013transition}. Graduate admissions are distinctive in their broadly shared timelines, with applicants navigating overlapping deadlines and admissions decisions. During such transitions, individuals often turn to online communities for information, advice, and support~\cite{gong2021smartest,deandrea2012serious}. In communities such as \texttt{r/gradadmissions} and \texttt{r/MBA}, applicants encounter not only practical guidance but also peers' experiences of uncertainty, setbacks, and distress.

Online communities also facilitate self-disclosure, allowing individuals to express emotions, make sense of their experiences, and seek validation~\cite{de2014mental,andalibi2016understanding,bazarova2014self}. Prior research has examined how supportive responses relate to recipients' subsequent mental-health risk~\cite{de2017language}, while engagement with supportive content, such as coping stories, has been associated with lower subsequent stress and depression~\cite{yuan2023mental}. However, the psychosocial implications of distress disclosures may extend beyond those sharing their experiences or receiving support~\cite{kim2023supporters}. Stress crossover theory suggests that stress-related experiences can extend between individuals through direct empathic responses or indirect interpersonal processes, while shared external stressors may also produce similar outcomes~\cite{westman2001stress}. This perspective highlights the importance of considering not only those who disclose distress but also others who encounter it.

In graduate-admissions communities, disclosures of rejection, academic qualifications, and financial uncertainty occur within a competitive application process. Encountering such disclosures may elicit empathic responses or unfavorable social comparisons, potentially heightening distress~\cite{vogel2014social,kramer2014experimental}. These possibilities are particularly relevant during application preparation, when applicants can actively shape their applications before entering stages increasingly dependent on external evaluations. Thus, engagement with peers' distress during preparation may be associated with psychosocial expression and participation during later admissions stages.

Understanding these associations is challenging because shared admissions deadlines and decisions may independently influence applicants' distress-related expression. Individuals experiencing similar concerns may also be more likely to interact, making interpersonal influence difficult to distinguish from pre-existing similarities and shared circumstances~\cite{shalizi2011homophily}. This raises a further question: \textit{whether psychosocial associations surrounding peer distress remain confined to directly exposed applicants or are also observable among others who subsequently interact with them.}

To address this gap, our work asks the following research questions (RQs):




\para{RQ1:} How is engagement with peers' distress during application preparation associated with users' psychosocial outcomes during the subsequent admissions stages?

\para{RQ2:} Are similar psychosocial associations observable among users who subsequently interact with previously exposed individuals, relative to those interacting with comparison users?

To answer our RQs, we conduct a large quasi-experimental study drawing on the potential outcomes framework~\cite{rubin2005causal}.
We focus on online communities supporting individuals through graduate admissions, a consequential transition marked by recurring application cycles and uncertainty about future plans. 
We analyze 171,981 posts and 1,130,413 comments from \texttt{r/gradadmissions} and \texttt{r/MBA} across three admissions cycles, spanning September 2022 through May 2025. 
Focusing on 32,301 posts published during the September–November application period, we identify exposure to admissions-related distress. 
We use stratified propensity-score matching on pre-exposure language, activity, and classifier-based distress signals to construct comparable groups of exposed and unexposed users. We then examine subsequent changes in behaviors and language.

We find that users who engaged with peers' distress during application preparation subsequently exhibited higher classifier-estimated anxiety, depression, and stress expression, alongside greater commenting activity and lexical redundancy in both communities. Increased participation was concentrated in commenting rather than original post creation, indicating that greater distress-related expression can coexist with continued community engagement. These associations recurred across admissions cycles, while no significant cycle heterogeneity was detected after correction. In contrast, we found no statistically significant associations among downstream users who subsequently interacted with previously exposed individuals.

Our work makes three contributions. First, we provide longitudinal empirical evidence of associations between engagement with peer distress and subsequent psychosocial expression and participation during a high-stakes life transition, informed by stress crossover theory. Second, we examine the extent to which these associations are observable beyond direct engagement, distinguishing exposure to distress disclosures from subsequent interactions with previously exposed users. Third, we offer design implications for online support communities, highlighting the need to preserve opportunities for distress-related self-disclosure and peer support while considering the psychosocial experiences of those encountering such content.

\section{Related Work}

\subsection{Distress Across Life Transitions and Temporally Structured Online Communities}

Online communities provide spaces for individuals to disclose psychological distress and seek peer support. Prior research has examined mental-health self-disclosure in dedicated support communities~\cite{de2014mental} and changes in anxiety-related concerns across phases of crises~\cite{chen2023you}, establishing the value of social-media language for understanding distress-related experiences and their temporal dynamics.

Beyond periods of crisis, prior work has examined psychosocial experiences during educational and life transitions. \citet{gong2021smartest} investigated stress and support needs during the transition out of college, while \citet{saha2025mental} examined changes in mental-health expression across stages of the COVID-19 pandemic in college communities. The timing of participation matters as well: users differ in how long they remain active~\cite{yang2010activity}, and cohorts entering during major external events show distinct patterns of engagement and retention~\cite{veselovsky2023reddit}.

The intersection of psychosocial expression, temporal context, and participation dynamics remains understudied in communities organized around recurring, high-stakes institutional transitions. Graduate admissions communities offer such a setting, where successive applicant cohorts navigate synchronized preparation, submission, and decision periods while individual participation may be concentrated within a single cycle.

\subsection{Stress Crossover in Online Interactions}

Stress crossover theory proposes that stress-related experiences may extend between individuals through direct empathic responses or indirect interpersonal processes, while shared stressors may also produce similar outcomes~\cite{westman2001stress}. Empirical work demonstrates stress responses to observing others' distress~\cite{engert2014cortisol}, while studies of online emotional contagion show associations between exposure to emotional content and subsequent expression~\cite{kramer2014experimental,ferrara2015measuring}. Research on online conversations and mental-health communities further highlights how emotional exchanges and the characteristics of peer support relate to psychosocial outcomes~\cite{kim2012you,saha2020causal}.

However, interpersonal influence remains difficult to distinguish from homophily and shared circumstances in observational networks~\cite{shalizi2011homophily}. Moreover, interacting with someone who has encountered distress does not necessarily entail encountering distress-related content. While prior work examines emotional exposure and interpersonal exchanges, less is understood about how associated psychosocial outcomes extend beyond direct engagement through subsequent community interactions, particularly during shared periods of institutional uncertainty.

\section{Data}
\paragraph{Communities.}
We study two Reddit communities centered on graduate-school admissions: \texttt{r/gradadmissions} and \texttt{r/MBA}. 
While \texttt{r/gradadmissions} hosts discussions of graduate applications across academic disciplines, \texttt{r/MBA} focuses on business-school admissions, including applicants transitioning from professional careers. Both communities provide settings in which prospective applicants discuss admissions uncertainty, academic qualifications, financial considerations, and application outcomes.

\paragraph{Data Source and Collection.}
We collected posts and comments from both communities using Arctic Shift ~\cite{heitmann2026arctic}. 
The corpus spans August~1, 2022 to July~30, 2025, covering three admissions cycles (2022--23, 2023--24, and 2024--25).

\paragraph{Cleaning and Filtering.}
After format standardization, we excluded deleted or removed records, known bots, comments containing fewer than five words, and content flagged as stylistically agent-like by a modified register classifier~\cite{goyal2026social}. 
The five-word threshold was applied to exclude contributions with minimal linguistic content, which can destabilize psycholinguistic measures and subsequent analyses, which can lead to instabilities~\cite{franklin-2015-theoretical}. The resulting corpus contains 1,302,394 contributions (171,981 posts and 1,130,413 comments) from 183,342 unique accounts (see \autoref{tab:cleaning-stage-flow}).

\begin{table}[t]
\centering
\small
\sffamily
\setlength{\tabcolsep}{4pt}
\begin{tabular}{lrrr}
Stage & Removed & \% in & Remaining \\
\midrule
\rowcolor{gray!10}Standardized corpus & --- & --- & $1{,}587{,}512$ \\
Deleted/removed & $81{,}835$ & $5.15$ & $1{,}505{,}677$ \\
\rowcolor{gray!10}Known bots & $3{,}623$ & $0.23$ & $1{,}502{,}054$ \\
$<5$ words & $193{,}872$ & $12.21$ & $1{,}308{,}182$ \\
\rowcolor{gray!10}Register filter & $5{,}788$ & $0.36$ & $1{,}302{,}394$ \\
\midrule
Retained & --- & $82.04$ & $1{,}302{,}394$ \\
\bottomrule
\end{tabular}
\caption{\textbf{Ordered filtration results.} Each removed record is attributed to the
first rule that excluded it.}
\label{tab:cleaning-stage-flow}
\end{table}

\para{Anchor Posts.}
Following prior work that identifies specific categories of social media content to examine psychosocial outcomes associated with exposure~\cite{saha2018social,yuan2023mental,pal2026hidden}, we define an \emph{anchor post} as an original post concerning a current, recent, or prospective graduate applicant or student that explicitly communicates a meaningful negative graduate-education signal. 
We identify seven qualifying categories: realized adverse outcomes; substantive negative affect; anticipated application threats; threatening social comparisons or vicarious threats; financial or affordability threats; career or degree-outcome threats; and serious life burdens, withdrawal, or abandonment. These signals may concern the author, other applicants, or broader admissions conditions, and do not require explicit personal distress. Routine profile evaluations, procedural questions, and neutral uncertainty alone do not qualify; however, posts containing such content are classified as anchors when they independently communicate at least one qualifying negative signal.

\paragraph{Anchor-Post Classification.}
We constructed a keyword-enriched, human-annotated benchmark of 400 original posts sampled across community--cycle groups from the September--November windows, with keywords used only to ensure balanced sampling, not to assign labels. The first author annotated each post against our operational definition, yielding 139 anchors and 261 non-anchors.

Anchor labels were produced by an LLM using a structured prompt that assessed applicant-timing scope, the seven qualifying negative-signal categories, and whether a post expressed only routine uncertainty; a post was labeled an anchor when it was in scope, met at least one category, and was not limited to routine uncertainty (prompt in Appendix~\ref{app:anchor_classification_prompt}). We evaluated four LLMs under stratified five-fold cross-validation (Table~\ref{tab:anchor-model-performance}) and selected Gemini~3.1 Flash-Lite, which achieved the highest recall (81.29\%) at competitive accuracy (83.00\%) and lower cost. We prioritized recall to capture a broader range of qualifying disclosures, accepting the precision trade-off.

Applying this classifier to 32,301 original posts (20,798 unique accounts) from the September--November windows identified 6,150 anchor posts (19.04\%).

\begin{table*}[t]
\centering
\small
\sffamily
\setlength{\tabcolsep}{6pt}
\begin{tabular}{lrrrrrr}
Model & Accuracy & Bal.\ Acc. & Precision & Recall & Spec. & $F_1$ \\
\midrule
\rowcolor{gray!10}Gemini 3.1 Flash-Lite & 83.00 & 82.60 & 72.90 & \textbf{81.29} & 83.91 & 76.87 \\
Gemini 3.1 Pro Preview & 80.50 & 73.79 & \textbf{86.75} & 51.80 & \textbf{95.79} & 64.86 \\
\rowcolor{gray!10}GPT-5.6 Terra & 83.75 & 81.16 & 78.91 & 72.66 & 89.66 & 75.66 \\
GPT-5.6 Luna & \textbf{85.00} & \textbf{83.46} & 78.42 & 78.42 & 88.51 & \textbf{78.42} \\
\bottomrule
\end{tabular}
\caption{Five-fold few-shot anchor-post classification on the 400-post
benchmark. All values except counts are percentages; Bal.\ Acc.\ is balanced
accuracy and Spec.\ is specificity.}
\label{tab:anchor-model-performance}
\end{table*}

\section{Methods}
To examine psychosocial outcomes associated with engagement with distress-related disclosures, we adopt a quasi-experimental design grounded in the potential outcomes framework~\cite{rubin2005causal}. We operationalize engagement as commenting on an anchor post during the September--November application period, using observable interaction as a proxy for encountering the post. Randomized assignment is infeasible in naturally occurring online communities, where controlling users' exposure to content would alter their usual interactions and raise ethical concerns~\cite{small2024protocols,moreno2013ethics}. Under the potential outcomes framework, each user has two potential outcomes, one under engagement ($Y_i(1)$) and one under non-engagement ($Y_i(0)$). Because both cannot be observed for the same user, we compare Treated and Control users with similar pre-engagement characteristics using propensity-score stratification~\cite{saha2019social,kiciman2018using,10.1145/3772318.3790476}, additionally adjusting for each user's baseline outcome.

\subsection{Operationalizing Psychosocial Outcomes}

Following prior computational social science research~\cite{saha2018social,saha2020psychosocial,pal2026algorithmic,yuan2026mental}, we operationalize psychosocial outcomes across three dimensions---affective, behavioral, and cognitive---comprising 17 primary outcomes (eight affective, five behavioral, four structural-linguistic). Each outcome is computed separately for a user's baseline and subsequent observation period, enabling comparisons that account for pre-existing differences in psychosocial expression and platform activity.

\paragraph{Affective Outcomes.}
We measure anxiety, depression, and stress-related linguistic expression using support-vector machine (SVM) classifiers following prior work~\cite{saha2019social}. For each user, we chronologically concatenate contributions separately within the baseline and outcome periods and apply each classifier to the resulting documents. We retain the continuous SVM decision margins rather than binary predictions, with higher scores indicating stronger alignment with the corresponding distress-related class. These scores are neither probabilities nor clinical assessments and are not comparable across classifiers. We additionally measure five LIWC-2015 categories~\cite{Pennebaker_Boyd_Jordan_Blackburn_2019}: positive emotion, negative emotion, anxiety, anger, and sadness, each as the percentage of matching words.

\paragraph{Behavioral Outcomes.}
We characterize participation through five measures of posting and commenting activity~\cite{guntuku2019language,saha2018social}: contribution, posting, and commenting frequency (counts per calendar day in the observation window); activity regularity (proportion of days with at least one contribution); and participation composition (proportion of contributions that are comments). Together these capture the frequency, regularity, and form of observable participation.

\paragraph{Cognitive Outcomes.}
Drawing on psycholinguistic research linking language use to psychological processes~\cite{tausczik2010psychological,ernala2017linguistic}, we examine four structural-linguistic measures: readability using Coleman--Liau index~\cite{liau1975computer}, verbosity (words per sentence), repeatability (lexical redundancy, $1-U/W$), and lexical complexity (letters per word), where $U$ and $W$ are unique lowercase word types and total word tokens. These characterize linguistic structure rather than directly assessing cognitive functioning. For each user and period, we concatenate all qualifying contributions before computing them.

\subsection{RQ1: Effect of Engagement}

\subsubsection{Engagement and Study Design}

We identify direct engagement by linking user comments to the 32,301 classified posts during September--November, the principal application-preparation period. Treated users are those who commented on an anchor post; Control users are those who commented on non-anchor posts without engaging with any anchor post in either study community during the same admissions cycle. Both groups required at least one prior contribution in their respective community, and we excluded users who authored anchor posts in either community. Each account contributes at most one observation per admissions cycle. These criteria, together with outcome availability and feasible placebo-date assignment, yielded 6,607 Treated and 3,826 Control user-cycle observations before matching.

Each admissions cycle spans August~1 through May~31 of the following year. The first qualifying anchor-post comment defines the index date for Treated users. Because Controls lack a natural engagement date, we assign placebo index dates within each community--cycle group to approximate the distribution of Treated users' index dates, minimizing differences in calendar timing while requiring eligible baseline activity and December--May outcomes. Baseline outcomes are measured from August~1 up to, but excluding, the index date. Contributions between the index date and December~1 are excluded as a post-engagement buffer. Outcomes are measured from December~1 through May~31, encompassing submissions, interviews, admission decisions, and waitlist outcomes; June--July activity is excluded.

\subsubsection{Analysis}

Users who engage with distress-related disclosures may differ systematically from those who do not in prior activity, linguistic expression, and discussion topics. To improve comparability, we construct, for each user-cycle, 518 baseline covariates from contributions across both study communities strictly before the index or placebo timestamp, spanning four families: activity and history measures, SVM-based anxiety/depression/stress scores, LIWC-2015 categories, and the 500 most frequent bigrams and trigrams ~\cite{saha2019social,Pennebaker_Boyd_Jordan_Blackburn_2019}. We require at least one pre-index contribution rather than a fixed minimum history length, retaining users with limited observed participation.

We use propensity-score stratification to identify Treated and Control users with similar observed pre-engagement characteristics. Following prior work~\cite{saha2019social,kiciman2018using}, we estimate each user's propensity from the 518 covariates using separate AdaBoost classifiers~\cite{hastie2009multi} per community--cycle group (200 depth-two trees, learning rate 0.05, fixed seed). We partition propensity scores into ten equal-width intervals, retaining strata with at least ten Treated and ten Control user-cycles. To target the average treatment effect on the treated (ATT), we assign unit weights to Treated users and within-stratum Treated-to-Control ratio weights to Controls. Of 33 populated strata, 23 met the retention criterion, yielding 8,903 user-cycle observations (5,100 Treated; 3,803 Control).

To assess balance, we compute standardized mean differences (SMDs) for all 518 covariates before and after stratification and weighting. The mean absolute SMD decreased from 0.0256 to 0.0224 and the maximum from 0.4129 to 0.1284; under a conservative 0.15 threshold, covariates exceeding it fell from seven to zero. These diagnostics indicate improved balance on measured characteristics, though they do not establish balance on unobserved factors.

For each of the 17 primary outcomes, we estimate an ATT-weighted, baseline-adjusted regression:
\begin{equation}
Y_{\mathrm{post}} = \alpha + \tau T + \beta Y_{\mathrm{baseline}} + \gamma_{\mathrm{stratum}} + \varepsilon,
\end{equation}
where $Y_{\mathrm{post}}$ is the December--May outcome, $T$ distinguishes Treated from Control users, $Y_{\mathrm{baseline}}$ is the pre-engagement value of the same outcome, and $\gamma_{\mathrm{stratum}}$ are propensity-stratum fixed effects. Together with ATT weighting, these adjustments compare subsequent outcomes among users with similar observed pre-engagement characteristics; the coefficient $\tau$ is the adjusted difference in December--May outcomes associated with earlier anchor-post engagement.

We estimate associations for the pooled sample and separately per community, clustering standard errors at the Reddit-account level to account for repeated observations across cycles. To compare across outcomes on different scales, we standardize each adjusted coefficient by the ATT-weighted standard deviation of the corresponding post-period outcome. We report nominal 95\% confidence intervals from clustered standard errors and assess significance using two-sided Wald tests, applying Benjamini--Hochberg false-discovery-rate correction separately within each outcome family, analysis scope, and model specification. We additionally test whether associations differ across cycles using engagement-by-cycle interactions and an omnibus Wald test.

\subsection{RQ2: Two-Hop Downstream Interaction}
\label{sec:methods-contagion}

\subsubsection{Engagement and Study Design}

To examine whether associations observed among directly engaged users extend through subsequent community interactions, we follow users from the RQ1 analysis. We identify contributions made by previously engaged users after their anchor-post interaction, and by comparison users after their assigned placebo dates, then identify downstream respondents who directly reply to these contributions. Respondents replying to previously engaged users form the Treated-chain group; those replying to comparison users form the Control-chain group.

We retain each downstream respondent's first qualifying interaction per admissions cycle and exclude respondents who themselves authored or commented on anchor posts during that cycle. We measure baseline outcomes from contributions strictly before this interaction and subsequent outcomes strictly afterward through May~31, requiring activity in both periods. Unlike RQ1, which measures outcomes during a fixed subsequent stage, RQ2 uses contact-relative observation windows.

\subsubsection{Analysis}

Although previously engaged and comparison users were balanced on observed baseline characteristics in RQ1, their downstream respondents may differ in prior activity, linguistic expression, and discussion topics, which may influence both whom they reply to and their subsequent outcomes. We therefore construct the same 518 pre-contact covariates and perform a second propensity-score analysis. Within each community--cycle group, AdaBoost classifiers estimate the propensity for a respondent's first eligible interaction to involve a previously engaged user rather than a comparison user---the likelihood of belonging to a Treated versus Control chain among users who have already made an eligible reply, not the likelihood of replying at all.

Following the RQ1 procedure, we divide propensity scores into ten equal-width strata, retaining those with at least ten downstream respondents per group, and weight Control-chain respondents to match the weighted Treated-chain mass within each stratum while incorporating the ATT weights inherited from RQ1. Of 33,191 identified downstream user-cycles, 19,105 had eligible pre- and post-contact activity, and 18,923 were retained after stratification (13,592 Treated-chain; 5,331 Control-chain). The mean absolute SMD decreased from 0.022 to 0.016 and the maximum from 0.209 to 0.059, with no covariates exceeding 0.15 after matching.

For each of the 17 primary outcomes, we estimate a weighted, baseline-adjusted regression comparing downstream respondents who replied to previously engaged users against those who replied to comparison users:
\begin{equation}
Y_{\mathrm{post}} = \alpha + \tau Z + \beta Y_{\mathrm{baseline}} + \gamma_{\mathrm{contact}} + \varepsilon,
\end{equation}
where $Z$ indicates whether the respondent's first qualifying reply was directed to a previously engaged user, $Y_{\mathrm{baseline}}$ is the pre-contact outcome, and $\gamma_{\mathrm{contact}}$ are community-by-contact-date fixed effects. The coefficient $\tau$ estimates the adjusted difference in subsequent outcomes between Treated-chain and Control-chain respondents. Contact-date adjustment accounts for shared calendar events, while clustering standard errors by the RQ1 user-cycle being replied to accounts for multiple respondents interacting with the same person. We estimate pooled, community-specific, and cycle-specific associations, reporting standardized adjusted effects and nominal 95\% confidence intervals, with significance assessed via two-sided Wald tests and Benjamini--Hochberg correction within each outcome family, analysis scope, and model specification.

\section{Results}

\subsection{RQ1: Psychosocial Outcomes Following Anchor-Post Engagement}

\autoref{tab:rq1-outcomes} presents standardized, baseline-adjusted associations between anchor-post engagement during application preparation and subsequent psychosocial outcomes. Positive coefficients indicate higher adjusted values among Treated users; negative, lower. Asterisks denote associations statistically significant after Benjamini--Hochberg correction ($p_{\mathrm{BH}} < .05$). Across both communities, we observe significant associations in distress-related expression, participation, and structural-linguistic patterns, examined by dimension below.
\begin{table}[t]
\centering
\sffamily
\footnotesize
\setlength{\tabcolsep}{3pt}
\resizebox{\columnwidth}{!}{%
\begin{tabular}{lcc}
Outcome & r/MBA & r/gradadmissions \\
\midrule
\rowcolor{blue!10}\multicolumn{3}{l}{\textit{Affective outcomes}} \\
\textbf{Anxiety SVM}
    & 0.244 [0.187, 0.301]*
    & 0.128 [0.042, 0.214]* \\
\textbf{Depression SVM}
    & 0.269 [0.213, 0.325]*
    & 0.141 [0.057, 0.225]* \\
\textbf{Stress SVM}
    & 0.256 [0.201, 0.310]*
    & 0.139 [0.058, 0.220]* \\
LIWC positive emotion
    & -0.001 [-0.063, 0.060]
    & 0.022 [-0.054, 0.097] \\
LIWC negative emotion
    & 0.067 [0.001, 0.133]
    & 0.067 [-0.011, 0.144] \\
LIWC anxiety
    & 0.013 [-0.048, 0.073]
    & -0.024 [-0.114, 0.066] \\
LIWC anger
    & 0.018 [-0.061, 0.097]
    & 0.048 [-0.026, 0.122] \\
LIWC sadness
    & 0.053 [-0.002, 0.108]
    & 0.007 [-0.070, 0.084] \\
\midrule
\rowcolor{blue!10}\multicolumn{3}{l}{\textit{Behavioral outcomes}} \\
\textbf{Contributions/day}
    & 0.325 [0.268, 0.383]*
    & 0.230 [0.156, 0.303]* \\
Posts/day
    & -0.024 [-0.086, 0.039]
    & -0.008 [-0.087, 0.071] \\
\textbf{Comments/day}
    & 0.331 [0.274, 0.388]*
    & 0.234 [0.161, 0.307]* \\
\textbf{Active-day share}
    & 0.432 [0.375, 0.488]*
    & 0.325 [0.247, 0.402]* \\
\textbf{Comment share}
    & 0.145 [0.082, 0.209]*
    & 0.201 [0.127, 0.275]* \\
\midrule
\rowcolor{blue!10}\multicolumn{3}{l}{\textit{Structural-linguistic outcomes}} \\
Readability
    & 0.053 [-0.012, 0.119]
    & 0.092 [0.011, 0.173]* \\
Verbosity
    & 0.038 [-0.022, 0.099]
    & 0.104 [0.022, 0.186]* \\
\textbf{Repeatability}
    & 0.336 [0.278, 0.393]*
    & 0.262 [0.187, 0.338]* \\
Complexity
    & 0.005 [-0.062, 0.071]
    & 0.019 [-0.062, 0.099] \\
\bottomrule
\end{tabular}%
}
\caption{Adjusted psychosocial outcomes following
anchor-post exposure across three admissions cycles ($^*$ indicates
Benjamini--Hochberg-adjusted $p<.05$.).}
\label{tab:rq1-outcomes}
\end{table}


\para{Affective Outcomes.}
Treated users exhibited significantly higher classifier-estimated anxiety, depression, and stress expression in both communities. In r/MBA, associations were strongest for depression ($\hat{\tau}_{\mathrm{std}}=0.269$), then stress ($0.256$) and anxiety ($0.244$); r/gradadmissions showed the same ordering at smaller magnitudes (depression: $0.141$; stress: $0.139$; anxiety: $0.128$). All six remained significant after correction, indicating that users who engaged with distress-related disclosures during preparation subsequently expressed more distress-related language than users with comparable pre-engagement characteristics. On the other hand, none of the five LIWC affective categories showed significant associations. Thus, the observed differences were concentrated in classifier-estimated distress expression rather than emotion-word frequencies.

\para{Behavioral Outcomes.} Treated users exhibited significantly greater subsequent participation in both communities, particularly through commenting. In r/MBA, commenting frequency ($\hat{\tau}_{\mathrm{std}}=0.331$), contribution frequency ($0.325$), and active-day share ($0.432$) were higher, with corresponding associations in r/gradadmissions ($0.234$, $0.230$, $0.325$). Comment share was also higher in both communities (r/MBA: $0.145$; r/gradadmissions: $0.201$), while posting frequency showed no significant association. Greater participation was therefore concentrated in ongoing conversations rather than new posts, with increased commenting accompanying higher distress-related classifier scores.

\para{Cognitive Outcomes.}
Treated users exhibited significantly greater lexical repeatability in both r/MBA ($\hat{\tau}_{\mathrm{std}}=0.336$) and r/gradadmissions ($0.262$), indicating greater lexical redundancy relative to comparable Controls. In r/gradadmissions, Treated users additionally showed higher verbosity ($0.104$) and Coleman--Liau reading-grade estimates ($0.092$), reflecting longer sentences and greater estimated textual difficulty; neither association was significant in r/MBA.

Across individual admissions cycles, estimates varied but no primary outcome showed statistically significant cycle variation after correction (Appendix \ref{app:rq1-cycle-heterogeneity}).
\subsection{RQ2: Downstream Psychosocial Outcomes}

\begin{table}[t]
\centering
\footnotesize
\sffamily
\setlength{\tabcolsep}{2pt}

\resizebox{\columnwidth}{!}{%
\begin{tabular}{@{}lcc@{}}
Outcome & r/MBA & r/gradadmissions \\
\midrule

\rowcolor{blue!10}\multicolumn{3}{@{}l}{\textit{Affective outcomes}} \\

Anxiety SVM
    & -0.011 [-0.065, 0.042]
    & 0.014 [-0.035, 0.062] \\
Depression SVM
    & 0.011 [-0.035, 0.057]
    & 0.016 [-0.034, 0.065] \\
Stress SVM
    & -0.005 [-0.051, 0.040]
    & 0.002 [-0.047, 0.050] \\
LIWC positive emotion
    & -0.004 [-0.057, 0.049]
    & 0.011 [-0.039, 0.062] \\
LIWC negative emotion
    & -0.003 [-0.054, 0.048]
    & 0.001 [-0.055, 0.058] \\
LIWC anxiety
    & -0.040 [-0.099, 0.018]
    & 0.016 [-0.032, 0.063] \\
LIWC anger
    & -0.011 [-0.063, 0.042]
    & 0.033 [-0.016, 0.083] \\
LIWC sadness
    & 0.008 [-0.041, 0.058]
    & 0.009 [-0.050, 0.067] \\

\midrule
\rowcolor{blue!10}\multicolumn{3}{@{}l}{\textit{Behavioral outcomes}} \\

Contributions/day
    & 0.010 [-0.029, 0.049]
    & -0.019 [-0.062, 0.024] \\
Posts/day
    & 0.033 [-0.007, 0.074]
    & -0.048 [-0.097, 0.000] \\
Comments/day
    & 0.009 [-0.029, 0.048]
    & -0.015 [-0.056, 0.026] \\
Active-day share
    & 0.015 [-0.022, 0.052]
    & 0.008 [-0.038, 0.054] \\
Comment share
    & 0.005 [-0.043, 0.053]
    & -0.001 [-0.048, 0.045] \\

\midrule
\rowcolor{blue!10}\multicolumn{3}{@{}l}{\textit{Structural-linguistic outcomes}} \\

Readability
    & 0.023 [-0.026, 0.072]
    & -0.020 [-0.065, 0.024] \\
Verbosity
    & 0.000 [-0.051, 0.051]
    & 0.007 [-0.039, 0.053] \\
Repeatability
    & 0.014 [-0.035, 0.063]
    & 0.023 [-0.026, 0.072] \\
Complexity
    & 0.013 [-0.034, 0.060]
    & -0.018 [-0.061, 0.025] \\

\bottomrule
\end{tabular}%
}

\caption{Adjusted associations between downstream contact
with previously exposed users and subsequent psychosocial
outcomes, pooled across admissions cycles.}
\label{tab:rq2-outcomes}
\end{table}

Across both communities, we found no statistically significant associations among downstream respondents for any of the 17 primary outcomes after Benjamini--Hochberg correction. This includes classifier-estimated distress expression, participation, and structural-linguistic measures, contrasting with the associations observed among directly exposed users in RQ1. These findings indicate that the psychosocial associations observed following direct engagement with distress disclosures were not similarly detectable among users who subsequently interacted with previously exposed individuals.

\section{Discussion}

\subsection{Peer Distress, Stress Crossover, and Continued Participation}

Viewed through stress crossover theory~\cite{westman2001stress}, engagement with peer distress during application preparation was associated with greater subsequent distress-related expression, potentially reflecting empathic responses or unfavorable social comparisons~\cite{vogel2014social}. Yet these users also commented more frequently and participated across more days. This raises the possibility that applicants continue seeking connection even while expressing greater distress. Thus, community participation may offer opportunities for support without necessarily indicating relief from admissions-related concerns.

This tension has implications for platform design. While online communities should preserve opportunities for self-disclosure and peer support, they must also consider the experiences of those encountering distress-related content. User-controlled exposure preferences, contextual framing of sensitive discussions, and stage-specific support resources offer possible ways to accommodate both needs without discouraging disclosure.

\subsection{The Extent of Observable Stress Crossover}

Although stress crossover theory proposes indirect interpersonal pathways~\cite{westman2001stress}, we found no significant psychosocial associations among downstream respondents. Unlike direct engagement with distress disclosures, subsequent contact with previously exposed users does not necessarily involve encountering distress-related content. Moreover, shared admissions pressures may independently shape users' distress-related expression~\cite{shalizi2011homophily}. Together, these findings suggest that understanding how far stress-related associations extend through online communities requires examining the content and context of interactions, rather than social connections alone.
\section{Conclusion}
Across three admissions cycles in r/MBA and r/gradadmissions, engagement with peer distress during application preparation was associated with greater subsequent distress-related linguistic expression, commenting activity, and lexical redundancy. These findings demonstrate that distress-related expression can coexist with continued participation during high-stakes life transitions. However, no significant associations were detected among downstream respondents, highlighting the distinction between direct exposure to distress disclosures and contact with previously exposed individuals. Together, our findings emphasize the importance of interaction content and shared contextual pressures when interpreting psychosocial experiences and social influence in online communities.

\section*{Limitations}

Our study has several limitations. First, although propensity-score stratification and baseline adjustments account for observed differences, the observational design cannot eliminate unmeasured confounding, self-selection, or shared admissions-related pressures. Second, anchor-post identification relies on an LLM classifier evaluated against a single-researcher-annotated, keyword-enriched benchmark of 400 posts. The absence of independent annotations, potential differences between the benchmark and broader corpus, and recall-oriented classifier selection introduce uncertainty and possible exposure misclassification. Third, commenting serves as a proxy for exposure but does not capture passive reading or users' interpretations of distress-related content. Our linguistic measures also capture distress-related expression rather than clinical mental-health states, while differences in text volume may influence SVM classifier scores and lexical repeatability. Furthermore, downstream contact does not guarantee exposure to distress-related content, and nonsignificant RQ2 findings do not rule out transmission. Finally, fixed admissions-stage windows simplify individual timelines, and findings from two Reddit communities may not generalize to other applicant populations or life transitions.

Finally, graduate and MBA admissions involve multiple application deadlines and rounds throughout the year. Our fixed analytical windows approximate a broader admissions cycle rather than individual applicants' timelines, potentially obscuring variation in when applicants experience preparation, submission, and decision-making. Findings from two Reddit communities may also not generalize to other applicant populations or life transitions.

\section*{Ethical Considerations}

This study examines publicly accessible Reddit discussions about graduate admissions, including potentially sensitive disclosures of distress, uncertainty, and personal setbacks. Although these discussions are publicly available, we recognize the importance of protecting users' privacy and minimizing risks of identification or unwanted scrutiny. Our analysis focuses on aggregate patterns of linguistic expression and community participation rather than evaluating individual users.

The use of automated distress classification also introduces risks of misclassification and misinterpretation. Our linguistic measures reflect distress-related expression rather than clinical mental-health states, and observed associations should not be interpreted as evidence that encountering peer distress causes psychological harm, nor as establishing stress transmission. We also recognize that these findings may not generalize equally across different applicant populations or linguistic backgrounds.

Finally, our findings should not be used to justify suppressing distress disclosures, identifying vulnerable individuals without their consent, or implementing automated mental-health surveillance. While our work offers insights for designing online communities that are sensitive to users encountering distress-related content, such efforts must preserve opportunities for self-disclosure and peer support. Potential interventions, including personalized content ranking or exposure controls, would require careful consideration of privacy, user autonomy, fairness, and unintended harms.

\section*{AI Involvement Disclosure}
 
The research we presented was conducted without the use of generative artificial intelligence tools for study design, data collection, analysis, implementation or the development of scientific contributions. Our use of language-editing tools (e.g., Grammarly, ChatGPT) was limited to improving the grammar and readability of certain sections. All scientific content, interpretations, and decisions reflect the original work, judgment and intellectual contributions of the research team.

\section*{Acknowledgment}
This work was supported in part by the Google Cloud Research Credits Program and the OpenAI Researcher Access Program.

\bibliography{references}
\clearpage

\appendix
\section{Appendix}
\label{sec:Appendix}

\label{app:data}

\subsection{Data Collection}

We collected original posts and comments from \texttt{r/MBA} and
\texttt{r/gradadmissions}. The \texttt{r/MBA} data span August 1, 2022 through
May 31, 2025, while the \texttt{r/gradadmissions} data span August 1, 2022
through July 30, 2025. We organized the records into three admissions cycles:
2022--23, 2023--24, and 2024--25, each beginning in August and ending the
following July.

The full combined corpus contained 1,587,512 records: 201,138 original posts
and 1,386,374 comments. Of these, 745,269 records came from \texttt{r/MBA}
and 842,243 came from \texttt{r/gradadmissions}. Table~\ref{tab:app-raw-corpus}
reports the corpus composition before content filtration.

\begin{table}[ht]
\centering
\small
\setlength{\tabcolsep}{3pt}
\begin{tabularx}{\columnwidth}{@{}Xrrr@{}}
\toprule
Community & Posts & Comments & Total \\
\midrule
\texttt{r/MBA}
    & 71,110 & 674,159 & 745,269 \\
\texttt{r/gradadmissions}
    & 130,028 & 712,215 & 842,243 \\
\midrule
Total
    & 201,138 & 1,386,374 & 1,587,512 \\
\bottomrule
\end{tabularx}
\caption{Corpus composition before content filtration.}
\label{tab:app-raw-corpus}
\end{table}

\subsection{Schema Standardization}

We mapped the original exports into a common post-and-comment schema. The
standardized fields included the community and record type; Reddit record,
post, and parent identifiers; author; Unix and ISO-8601 timestamps; post title
and body or comment body; Reddit score; admissions cycle; and source-file
provenance. Post titles and bodies were joined using two newline characters,
whereas comment bodies were used directly.

Standardization did not lowercase text, remove punctuation or emoji, normalize
whitespace, or otherwise alter the original language. Standardization changed the representation of the records but did not
remove any observations.

\subsection{Content Filtration}

We applied four exclusion rules sequentially. First, we removed records whose
author or text was marked as deleted or removed by Reddit. Second, we removed
records produced by known automated accounts or containing an explicit bot
self-identification phrase. Third, we removed records containing fewer than
five whitespace-delimited words. Finally, we applied the modified register Moltbook-style classifier and removed records with an agent-style probability
of at least 0.75.

The filtration stages were applied in a fixed order, and each excluded record
was assigned to the first rule that removed it. Text normalization used by the
agent-style classifier was temporary and did not alter the text retained for
analysis.

\begin{table}[ht]
\centering
\footnotesize
\setlength{\tabcolsep}{3pt}
\begin{tabularx}{\columnwidth}{@{}Xrrr@{}}
\toprule
Stage & Removed & \% corpus & Remaining \\
\midrule
Standardized corpus
    & --- & --- & 1,587,512 \\
Deleted or removed records
    & 81,835 & 5.15 & 1,505,677 \\
Known bots or bot phrases
    & 3,623 & 0.23 & 1,502,054 \\
Fewer than five words
    & 193,872 & 12.21 & 1,308,182 \\
Agent-style score $\geq 0.75$
    & 5,788 & 0.36 & 1,302,394 \\
\midrule
Retained corpus
    & --- & 82.04 & 1,302,394 \\
\bottomrule
\end{tabularx}
\caption{Sequential content-filtration results. Percentages use the
standardized corpus as the denominator.}
\label{tab:app-filtering}
\end{table}

The final cleaned corpus contained 1,302,394 records: 171,981 original posts
and 1,130,413 comments. This included 631,497 records from \texttt{r/MBA} and
670,897 records from \texttt{r/gradadmissions}.

\begin{table}[ht]
\centering
\small
\setlength{\tabcolsep}{3pt}
\begin{tabularx}{\columnwidth}{@{}Xrrr@{}}
\toprule
Community & Posts & Comments & Total \\
\midrule
\texttt{r/MBA}
    & 60,707 & 570,790 & 631,497 \\
\texttt{r/gradadmissions}
    & 111,274 & 559,623 & 670,897 \\
\midrule
Total
    & 171,981 & 1,130,413 & 1,302,394 \\
\bottomrule
\end{tabularx}
\caption{Retained records by community and record type.}
\label{tab:app-cleaned-corpus}
\end{table}

\subsection{Study Windows}

We used different portions of each admissions cycle for exposure identification
and outcome measurement. Anchor-post classification and exposure identification
used posts and comments made from September 1 through November 30. Baseline
activity consisted of all observed same-cycle contributions strictly before a
user's treatment or placebo index timestamp. All primary RQ1 behavioral,
affective, and structural-linguistic outcomes were measured using activity from
December 1 through May 31.

For RQ2, a downstream user's baseline consisted of all observed same-cycle
activity strictly before their first eligible contact with a focal user.
Post-contact outcomes consisted of activity strictly after that contact through
May 31 of the same admissions cycle.

\subsection{September--November Post Corpus}

The anchor-post classification corpus contained 32,301 cleaned original posts
published during September--November across the three admissions cycles.
It included 13,399 posts from \texttt{r/MBA} and 18,902 posts from
\texttt{r/gradadmissions}. The final labels identified 6,150 anchor posts
(19.04\%) and 26,151 non-anchor posts (80.96\%).

\begin{table}[ht]
\centering
\footnotesize
\setlength{\tabcolsep}{2pt}
\begin{tabularx}{\columnwidth}{@{}Xrrrr@{}}
\toprule
Community & Posts & Anchors & Non-anchor & Anchor \% \\
\midrule
\texttt{r/MBA}
    & 13,399 & 2,383 & 11,016 & 17.78 \\
\texttt{r/gradadmissions}
    & 18,902 & 3,767 & 15,135 & 19.93 \\
\midrule
Total
    & 32,301 & 6,150 & 26,151 & 19.04 \\
\bottomrule
\end{tabularx}
\caption{Final September--November post labels by community.}
\label{tab:app-labelled-posts}
\end{table}

\begin{table}[ht]
\centering
\footnotesize
\setlength{\tabcolsep}{2pt}
\begin{tabularx}{\columnwidth}{@{}Xrrrr@{}}
\toprule
Cycle & Posts & Anchors & Non-anchor & Anchor \% \\
\midrule
2022--23
    & 6,784 & 1,112 & 5,672 & 16.39 \\
2023--24
    & 12,122 & 2,367 & 9,755 & 19.53 \\
2024--25
    & 13,395 & 2,671 & 10,724 & 19.94 \\
\midrule
Total
    & 32,301 & 6,150 & 26,151 & 19.04 \\
\bottomrule
\end{tabularx}
\caption{Final September--November post labels by admissions cycle.}
\label{tab:app-labelled-posts-cycle}
\end{table}

\section{Graduate-Admissions Anchor Classification Prompt}
\label{app:anchor_classification_prompt}

The following reproduces the complete instruction prompt used to classify original posts from graduate-admissions communities as \texttt{anchor} or \texttt{non\_anchor}. The prompt includes the scope definition, qualifying signals, decision safeguards, evidence-extraction rules, and required output schema.

\begin{promptbox}{Graduate-admissions anchor classification: full prompt}
You are classifying original Reddit posts from graduate-admissions communities.\\
\par\smallskip
Choose exactly one final label:\\
\par\smallskip
- anchor\\
- non\_\allowbreak{}anchor\\
\par\smallskip
\textbf{PURPOSE}\par
\par\smallskip
Identify negative graduate-education discourse: observable adversity, negative\\
affect, threatening comparison, consequential uncertainty, or plausible\\
negative outcomes connected to applying to, financing, attending, completing,\\
or obtaining the expected outcome from graduate education.\\
\par\smallskip
The construct includes personal disclosures, social comparison, vicarious\\
threat, collective threats, and negative information environments.\\
\par\smallskip
Do not infer a clinical diagnosis, psychological deterioration, or the actual\\
effect on a reader. Classify only what the post itself communicates. Do not\\
decide whether the author's concern is objectively justified or easily solved.\\
\par\smallskip
\textbf{APPLICANT-TIMING SCOPE}\par
\par\smallskip
The post must concern a current applicant, reapplicant, recent or past\\
applicant, current graduate student, admitted student, or a person making an\\
immediate graduate-education application, enrollment, financing, completion,\\
or degree-outcome decision.\\
\par\smallskip
Classify as non\_\allowbreak{}anchor when the author explicitly identifies themselves as an\\
early-stage undergraduate or otherwise states that they are not yet applying\\
and discusses only a distant future application. A junior asking how a present\\
choice might affect an MBA application years later is outside scope.\\
\par\smallskip
\textbf{BINARY DECISION RULE}\par
\par\smallskip
Classify a post as anchor when both conditions are satisfied:\\
\par\smallskip
1. The post is within the graduate-education and applicant-timing scope.\\
\par\smallskip
2. The post directly communicates at least one qualifying negative signal\\
   described below.\\
\par\smallskip
The qualifying categories are alternatives. Evidence satisfying any one\\
category can qualify. Substantive negative affect does not also require a\\
realized adverse outcome, and a realized or collective adverse outcome does\\
not require explicit emotional language.\\
\par\smallskip
\textbf{QUALIFYING NEGATIVE SIGNALS}\par
\par\smallskip
\textbf{A. REALIZED OR SALIENT ADVERSE OUTCOME}\par
\par\smallskip
Examples include:\\
\par\smallskip
- rejection or an unsuccessful application cycle;\\
- consequential waitlisting or adverse deferral;\\
- an application withdrawn, closed, or not processed in a consequential way;\\
- funding loss or inability to attend;\\
- failure to obtain an expected educational or career outcome;\\
- admissions difficulty that causes reapplication, delay, remediation, or a\\
  compensatory work or research year;\\
- reports or questions about adverse outcomes experienced by other applicants.\\
\par\smallskip
A post may qualify when it reports, compares, solicits, or interprets rejection\\
or other adverse outcomes experienced by others. The author does not need to\\
have personally experienced the outcome.\\
\par\smallskip
A bare historical reference to rejection does not qualify when it is merely\\
background for an otherwise neutral procedural question.\\
\par\smallskip
A temporary administrative hold or routine request for a missing document\\
does not qualify by itself when processing can resume after the requirement is\\
satisfied and the author neutrally asks how to comply. It may qualify when the\\
requirement cannot be satisfied, a deadline has passed, the application has\\
been closed, or the author frames it as likely to cause disqualification or\\
another consequential loss.\\
\par\smallskip
\textbf{B. EXPLICIT NEGATIVE AFFECT}\par
\par\smallskip
Examples include:\\
\par\smallskip
- fear, anxiety, hopelessness, devastation, discouragement, or emotional\\
  strain;\\
- feeling lost, overwhelmed, defeated, terrified, or unable to cope;\\
- explicit distress about an admissions, educational, financial, immigration,\\
  or career consequence;\\
- sustained burden, repeated or intensified emotion, difficulty disengaging,\\
  or a sense that continued effort will never be enough.\\
\par\smallskip
Substantive negative affect may independently qualify a post as anchor. It\\
does not also require rejection, likely failure, financial harm, or another\\
concrete adverse outcome.\\
\par\smallskip
Evaluate affect cumulatively. Several connected phrases may establish\\
substantive affect through duration, repetition, intensity, inability to\\
disengage, or the author's description of the burden.\\
\par\smallskip
However, mild expressions such as "worried," "concerned," "nervous,"\\
"confused," or "stressed" do not qualify by themselves when they merely\\
introduce an ordinary, actionable question.\\
\par\smallskip
Do not infer qualifying affect solely from someone requesting advice,\\
reassurance, or an evaluation. Conversely, do not dismiss substantive affect\\
merely because the post also requests strategic or procedural advice.\\
\par\smallskip
\textbf{C. EXPECTED FAILURE OR APPLICATION THREAT}\par
\par\smallskip
Examples include:\\
\par\smallskip
- belief or clearly expressed fear that rejection, disqualification, or\\
  failure is likely;\\
- belief that a fixed profile weakness or exhausted opportunity to improve may\\
  lead to rejection or serious application harm;\\
- belief that further effort will still be insufficient;\\
- a specific application mechanism explicitly connected to a consequential\\
  negative outcome such as rejection, disqualification, loss of\\
  consideration, or inability to enroll;\\
- belief that an application may not receive meaningful consideration;\\
- concern that every available choice carries meaningful application harm.\\
\par\smallskip
An anticipated or hypothetical threat may qualify when the post identifies a\\
specific mechanism and explicitly connects it to a consequential negative\\
outcome. The threat does not need to have already occurred, be objectively\\
verified, or be statistically likely.\\
\par\smallskip
A generic question about whether an ordinary choice, profile feature, or\\
application component might help or hurt does not qualify by itself. A routine\\
profile weakness does not qualify merely because the author asks how to\\
compensate for it.\\
\par\smallskip
A concrete eligibility rule, restriction, missing item, or procedural barrier\\
does not qualify by itself when it is neutrally presented as a factual or\\
compliance question. It qualifies when the post frames it as a consequential\\
setback, continuing jeopardy, feared rejection, inability to proceed, or\\
another meaningful threat.\\
\par\smallskip
\textbf{D. THREATENING COMPARISON OR VICARIOUS THREAT}\par
\par\smallskip
Examples include:\\
\par\smallskip
- competition portrayed as fierce or admission as nearly impossible;\\
- the author describing themselves as substantially disadvantaged compared\\
  with other applicants;\\
- strong applicants being rejected in a way that communicates threat to\\
  comparable applicants;\\
- widespread rejection, a "bloodbath," worsening competition, or normally\\
  successful applicants failing in a potentially recurring admissions cycle;\\
- reports that applicants or graduates like the author cannot obtain expected\\
  funding, admission, employment, or career outcomes;\\
- a strong applicant treating a comparatively minor weakness as potentially\\
  making the otherwise strong profile inadequate for competitive admission.\\
\par\smallskip
Personal distress is not required for a collective or vicarious threat.\\
\par\smallskip
Do not classify a neutral profile as anchor merely because a reader could\\
compare themselves with it. A strong profile, high score, low score, or past\\
mistake is not sufficient by itself. The post must itself frame the comparison\\
or overall profile through insecurity, feared inadequacy, rejection, group\\
failure, or another observable negative signal.\\
\par\smallskip
Use evidence inside the post whenever possible, including the author's own\\
positive description of an otherwise strong result followed by concern that a\\
minor weakness may negate it. Do not invent external admissions benchmarks.\\
\par\smallskip
\textbf{E. FUNDING, AFFORDABILITY, OR ROI THREAT}\par
\par\smallskip
Examples include:\\
\par\smallskip
- inability to attend without funding;\\
- unaffordable tuition or living costs;\\
- inadequate or lost funding;\\
- burdensome debt or significant opportunity cost;\\
- family or employment obligations that threaten financial feasibility;\\
- credible concern that the degree will not produce enough benefit to justify\\
  its cost;\\
- fear that pursuing the degree could create financial harm.\\
\par\smallskip
Financial information and ROI evaluation are not automatically threats.\\
\par\smallskip
Classify as non\_\allowbreak{}anchor when tuition, assistantships, living expenses, salary\\
targets, brand recognition, or ROI are used primarily as neutral criteria for\\
comparing viable programs or deciding whether a degree may be useful.\\
\par\smallskip
Classify as anchor when the post connects those considerations to inability to\\
attend, serious financial strain, burdensome debt, significant sacrifice,\\
feared financial harm, or belief that the degree will fail to provide a\\
necessary expected benefit.\\
\par\smallskip
\textbf{F. CAREER OR DEGREE-OUTCOME THREAT}\par
\par\smallskip
Examples include:\\
\par\smallskip
- poor recruiting, placement difficulty, unemployment, or lack of jobs;\\
- career stagnation, insecurity, or failed career transition;\\
- degree regret;\\
- concern that graduates cannot obtain expected opportunities;\\
- fear that a program could harm rather than advance a career;\\
- belief that attending a particular program will foreclose the author's\\
  intended career;\\
- questioning whether graduate education remains worthwhile under an adverse\\
  job market.\\
\par\smallskip
Career content qualifies only when the adverse condition is connected to\\
applying to, financing, attending, completing, or expecting an outcome from\\
graduate education.\\
\par\smallskip
Neutral questions about career goals, salary, employment statistics, regional\\
brand value, whether a master's might help, or whether experience might be\\
preferable are non\_\allowbreak{}anchor without an adverse career condition or feared degree\\
failure.\\
\par\smallskip
\textbf{G. LIFE BURDEN, WITHDRAWAL, OR ABANDONMENT}\par
\par\smallskip
Examples include:\\
\par\smallskip
- serious conflict involving graduate education and caregiving, family,\\
  employment, health, relocation, or immigration;\\
- concern that major obligations threaten the ability to attend or complete\\
  the degree;\\
- stopping applications;\\
- withdrawing an application;\\
- declining attendance because the path is untenable;\\
- leaving a program or giving up on graduate education.\\
\par\smallskip
Ordinary scheduling questions and neutral selection between opportunities do\\
not qualify.\\
\par\smallskip
\textbf{CONTROLLABILITY, CONSEQUENCE, AND FORMAT}\par
\par\smallskip
Consider whether the author presents the concern as reasonably addressable or\\
as a threat they may be unable to avert. A fixed circumstance, exhausted\\
opportunity to improve, or belief that further effort will still be\\
insufficient may qualify when connected to rejection, failure, or another\\
negative graduate-education outcome.\\
\par\smallskip
A request for an actionable strategy remains non\_\allowbreak{}anchor when the author does\\
not express expected failure, futility, substantive negative affect, or a\\
consequential negative outcome.\\
\par\smallskip
A qualifying signal does not need to dominate the entire post. One clear\\
clause may qualify when it independently communicates substantive adversity,\\
negative affect, threatening comparison, expected failure, or another\\
consequential threat.\\
\par\smallskip
A clause does not qualify when its negativity consists only of mild concern,\\
ordinary uncertainty, a routine tradeoff, a bare weakness, or asking whether\\
an ordinary choice might help or hurt.\\
\par\smallskip
The post's format does not determine the label. Advice-seeking, procedural,\\
profile-oriented, or strategic posts may be anchors when they contain an\\
independently qualifying signal. Those formats also do not become anchors\\
merely because they mention rejection, weakness, cost, competition, or jobs.\\
\par\smallskip
\textbf{ROUTINE-QUESTION SAFEGUARD}\par
\par\smallskip
When a post is primarily a routine profile, chances, test, recommender,\\
document, deadline, school-selection, program-comparison, career-strategy, or\\
application-strategy question, classify it as non\_\allowbreak{}anchor if its only negative\\
content is:\\
\par\smallskip
- a mild expression of worry or concern;\\
- a generic question about whether an ordinary choice could help or hurt;\\
- uncertainty about whether a profile component is sufficient;\\
- a request for reassurance about chances;\\
- a past weakness followed by recovery and a question about how to explain or\\
  compensate for it;\\
- a resolved procedural mistake or temporary administrative hold;\\
- neutral consideration of cost, ROI, brand, salary, or career tradeoffs.\\
\par\smallskip
Do not combine a mild concern phrase with a neutral advice request to create a\\
qualifying threat. Phrases such as "what should I do," "best course of action,"\\
"how did you get in," and "any advice" are neutral requests and cannot serve\\
as evidence of threat.\\
\par\smallskip
Classify such a post as anchor when it additionally communicates substantive\\
negative affect, a specific consequential threat, inability to avert harm,\\
threatening comparison, serious financial or career harm, realized adversity,\\
or another independently qualifying signal.\\
\par\smallskip
\textbf{IMPORTANT BOUNDARIES}\par
\par\smallskip
Never classify from an isolated keyword. Interpret each expression within the\\
complete phrase and post.\\
\par\smallskip
Words such as "rejected," "worried," "concerned," "chance," "competitive,"\\
"low," "weak," "risk," "worth," "afford," "debt," "bad," "appalling," and\\
"long shot" do not determine the label by themselves.\\
\par\smallskip
Resolve who might reject what. Fear that a professor may decline a request for\\
a recommendation is not automatically fear of graduate-admission rejection.\\
Humorous or softened discomfort such as "fear of rejection LOL" does not\\
qualify without another consequential signal.\\
\par\smallskip
Do not resolve the author's perceived threat for them. A reassuring fact, such\\
as a score being optional, does not erase directly expressed substantive\\
affect, perceived irreversibility, or feared application harm.\\
\par\smallskip
Do not treat a concern as factually substantiated only when it is objectively\\
correct. Classify the negative discourse communicated by the post.\\
\par\smallskip
"Deferred MBA" usually describes a program type and is not an adverse\\
deferral. Technical phrases such as "stress testing" do not indicate distress.\\
Positive uses such as "overwhelmed by the quality of the feedback" do not\\
indicate negative affect.\\
\par\smallskip
Negative language appearing only in quoted or copied material does not qualify\\
unless the author adopts the concern.\\
\par\smallskip
General discussion of a poor job market does not qualify unless a graduate\\
degree, application, enrollment decision, or expected degree outcome is\\
implicated. Conversely, a collective admissions or degree-outcome threat does\\
not require personal self-disclosure.\\
\par\smallskip
\textbf{EVIDENCE EXTRACTION}\par
\par\smallskip
Copy the shortest exact phrases supporting an anchor classification into\\
evidence\_\allowbreak{}phrases.\\
\par\smallskip
Each evidence phrase must independently support the category for which it is\\
cited. Do not combine an unrelated mild concern with a neutral request to\\
manufacture a qualifying signal.\\
\par\smallskip
For expected failure or application threat, the evidence must state or clearly\\
imply a consequential negative outcome. A neutral request such as "best course\\
of action" is not threat evidence.\\
\par\smallskip
Copy exact phrases supporting a neutral interpretation into\\
counterevidence\_\allowbreak{}phrases. Do not place negative-affect or threat evidence in\\
counterevidence merely because the post also asks a procedural question.\\
\par\smallskip
Do not extract isolated keywords when a longer phrase is necessary to preserve\\
meaning.\\
\par\smallskip
Every extracted phrase must appear verbatim in the target post. Do not invent,\\
paraphrase, correct, or add text. If a proposed phrase cannot be found exactly\\
in the target post, do not use it. Use an empty list when no relevant phrase\\
exists.\\
\par\smallskip
\textbf{DECISION PROCESS}\par
\par\smallskip
Before assigning the label:\\
\par\smallskip
1. Confirm graduate-education relevance and applicant-timing scope.\\
\par\smallskip
2. Identify the strongest exact phrase supporting each potentially applicable\\
   qualifying category.\\
\par\smallskip
3. Determine whether the phrase establishes substantive affect, adversity, a\\
   specific consequential threat, threatening comparison, or another\\
   qualifying signal rather than a neutral fact, routine tradeoff, generic\\
   possibility, or ordinary request.\\
\par\smallskip
4. Apply the routine-question safeguard. Do not allow mild concern, a bare\\
   weakness, a neutral comparison, or a solvable procedural question to create\\
   a qualifying category.\\
\par\smallskip
5. If an independently qualifying signal remains, do not override it merely\\
   because the post is also strategic, procedural, profile-oriented, or\\
   advice-seeking.\\
\par\smallskip
6. Check that every evidence phrase is copied verbatim and actually supports\\
   the selected category.\\
\par\smallskip
7. If no qualifying signal can be supported with an exact phrase, choose\\
   non\_\allowbreak{}anchor.\\
\par\smallskip
\textbf{FINAL CONSISTENCY RULE}\par
\par\smallskip
If graduate-education relevance and applicant-timing scope are satisfied and\\
at least one qualifying negative category is true, final\_\allowbreak{}label must be anchor.\\
\par\smallskip
If final\_\allowbreak{}label is non\_\allowbreak{}anchor, every qualifying negative category must be\\
false. The neutral-information or routine-uncertainty field may be true even\\
when a post also contains negative words.\\
\par\smallskip
Return exactly one valid JSON object using this schema:\\
\par\smallskip
\par\smallskip{\ttfamily\small
\mbox{}\{\\
\mbox{}  "in\_\allowbreak{}scope": true,\\
\mbox{}  "realized\_\allowbreak{}adverse\_\allowbreak{}outcome": false,\\
\mbox{}  "substantive\_\allowbreak{}negative\_\allowbreak{}affect": false,\\
\mbox{}  "consequential\_\allowbreak{}anticipated\_\allowbreak{}threat": false,\\
\mbox{}  "threatening\_\allowbreak{}comparison\_\allowbreak{}vicarious\_\allowbreak{}threat\_\allowbreak{}or\_\allowbreak{}negative\_\allowbreak{}information\_\allowbreak{}environment": false,\\
\mbox{}  "funding\_\allowbreak{}affordability\_\allowbreak{}or\_\allowbreak{}roi\_\allowbreak{}threat": false,\\
\mbox{}  "career\_\allowbreak{}or\_\allowbreak{}degree\_\allowbreak{}outcome\_\allowbreak{}threat": false,\\
\mbox{}  "life\_\allowbreak{}burden\_\allowbreak{}withdrawal\_\allowbreak{}or\_\allowbreak{}abandonment": false,\\
\mbox{}  "only\_\allowbreak{}routine\_\allowbreak{}uncertainty": true,\\
\mbox{}  "evidence\_\allowbreak{}phrases": [],\\
\mbox{}  "decision\_\allowbreak{}basis": "Brief explanation grounded only in the post.",\\
\mbox{}  "final\_\allowbreak{}label": "non\_\allowbreak{}anchor",\\
\mbox{}  "confidence": 0.90\\
\mbox{}\}\\
}\par\smallskip
\par\smallskip
Map the substantive categories defined above to the corresponding fields in\\
this schema. Set `in\_\allowbreak{}scope` to true only when both graduate-education relevance\\
and applicant-timing scope are satisfied.\\
\par\smallskip
Set `only\_\allowbreak{}routine\_\allowbreak{}uncertainty` to true only when no independently qualifying\\
negative category remains after applying the routine-question safeguard.\\
\par\smallskip
The final label must be `anchor` if and only if `in\_\allowbreak{}scope` is true, at least\\
one of the seven qualifying-category fields is true, and\\
`only\_\allowbreak{}routine\_\allowbreak{}uncertainty` is false. Otherwise it must be `non\_\allowbreak{}anchor`.\\
\par\smallskip
For `anchor`, `evidence\_\allowbreak{}phrases` must contain exact supporting text. For\\
`non\_\allowbreak{}anchor`, `evidence\_\allowbreak{}phrases` must be an empty list.\\
\par\smallskip
Return JSON only. Do not include Markdown or additional commentary.\\
\end{promptbox}

\section{Matching Implementation}
\label{app:matching}

\subsection{Pre-Index Covariates}

For each eligible user-cycle, we aggregated all nonempty posts and comments
observed in the two study communities during the corresponding admissions
cycle, strictly before the treatment or placebo index timestamp. The baseline
extended from the user's first observed same-cycle contribution to immediately
before the index event. Eligibility required at least one pre-index
contribution.

We computed 518 baseline covariates per user-cycle: six activity and
observation-history measures, three mental-health language scores, nine
LIWC-2015 proportions, and 500 lexical-content features
(Table~\ref{tab:app-matching-covariates}).

\begin{table}[t]
\centering
\small
\setlength{\tabcolsep}{4pt}
\begin{tabular}{@{}p{0.22\columnwidth}p{0.55\columnwidth}r@{}}
\toprule
Family & Measures & Count \\
\midrule
Activity and history &
Contribution rate, post share, active-day share, mean words per contribution,
focal-community share, and observed-history duration & 6 \\
Mental-health language &
Anxiety, depression, and stress SVM scores & 3 \\
Psycholinguistic &
Nine LIWC-2015 categories & 9 \\
Lexical content &
L1-normalized bigram and trigram frequencies & 500 \\
\midrule
Total & & 518 \\
\bottomrule
\end{tabular}
\caption{Baseline covariates used for propensity-score estimation.}
\label{tab:app-matching-covariates}
\end{table}

Contribution rate was the number of baseline posts and comments divided by
the number of days from the user's first observed baseline contribution
through the index date. Post share was the proportion of baseline
contributions that were posts, and active-day share was the proportion of
observed baseline days with at least one contribution. Mean words per
contribution was calculated across baseline records. Focal-community
contribution share measured the proportion of contributions in the community
defining the index interaction, while observed-history duration measured
the number of calendar days between the first baseline contribution and index.

For language-based covariates, all baseline contributions belonging to a
user-cycle were concatenated into one document. Anxiety, depression, and
stress classifiers were applied to this document, producing native,
uncalibrated SVM decision-function scores. LIWC covariates represented
the proportion of word tokens matching each selected dictionary category:
positive emotion, negative emotion, anxiety, anger, sadness, risk,
comparison, cognitive processes, and first-person singular language.

Lexical-content features comprised the 500 most frequent bigrams and
trigrams occurring in at least five baseline user-cycle documents.
Feature vectors were normalized to unit L1 frequency within each document.
No post-index activity or outcome variable entered propensity-score
estimation.

\subsection{RQ1 Propensity-Score Estimation}

We estimated the probability of anchor-post engagement separately within
each community and admissions cycle, producing six propensity-score models.
Each model used all 518 pre-index covariates and AdaBoost with 200 depth-two
decision-tree base learners, a learning rate of 0.05. A community-cycle stratum was retained
only if it contained at least ten treated and ten control user-cycles,
excluding regions with insufficient treated-control overlap.

Within each retained stratum, treated observations received weight one.
Control observations received an average-treatment-on-the-treated (ATT)
weight of

\begin{equation}
w_{0s} = \frac{N_{1s}}{N_{0s}},
\end{equation}

where $N_{1s}$ and $N_{0s}$ denote the treated and control counts in
retained stratum $s$. Thus, weighted control mass equaled treated mass
within each stratum.

Propensity estimation began with 10,433 eligible user-cycles, including
6,607 treated observations and 3,826 controls. Across the six community-cycle
models, 33 propensity-score strata contained observations, of which 23
satisfied the minimum group-size requirement. The final RQ1 analytic
cohort comprised 8,903 user-cycles: 5,100 treated observations and
3,803 controls (Table~\ref{tab:app-rq1-matching-flow}).

\begin{table}[ht]
\centering
\small
\setlength{\tabcolsep}{4pt}
\begin{tabular}{@{}lrrr@{}}
\toprule
Stage & Treated & Control & Total \\
\midrule
Eligible & 6,607 & 3,826 & 10,433 \\
Retained & 5,100 & 3,803 & 8,903 \\
\bottomrule
\end{tabular}
\caption{RQ1 propensity-score stratification and retention.}
\label{tab:app-rq1-matching-flow}
\end{table}

\subsection{RQ1 Covariate Balance}

We assessed covariate balance using the absolute standardized mean
difference (SMD) for each of the 518 baseline covariates:

\begin{equation}
\operatorname{SMD}(X) =
\frac{\bar{X}_{1}-\bar{X}_{0}}
{\sqrt{(S_{1}^{2}+S_{0}^{2})/2}}.
\end{equation}

Group means and variances were calculated without weights before matching
and using final ATT weights afterward. Diagnostics were computed for
the combined analytic cohort and separately within each community-cycle
model.

Across all 518 covariates, mean absolute SMD decreased from 0.0256 before
stratification to 0.0224 afterward. Maximum absolute SMD decreased from
0.4129 to 0.1284. Seven covariates exceeded an absolute SMD of 0.15
before matching, compared with none afterward.
Table~\ref{tab:app-balance-summary} reports balance diagnostics for
both RQ1 and RQ2.

\begin{table}[ht]
\centering
\small
\setlength{\tabcolsep}{4pt}
\begin{tabular}{@{}llrrr@{}}
\toprule
Analysis & Stage & Mean & Max. & $>0.15$ \\
\midrule
RQ1 & Before & 0.0256 & 0.4129 & 7 \\
RQ1 & After  & 0.0224 & 0.1284 & 0 \\
RQ2 & Before & 0.0225 & 0.2094 & 2 \\
RQ2 & After  & 0.0162 & 0.0594 & 0 \\
\bottomrule
\end{tabular}
\caption{Absolute standardized mean differences before and after matching.}
\label{tab:app-balance-summary}
\end{table}

In Table~\ref{tab:app-balance-summary}, mean and maximum values refer
to absolute SMDs, while the final column counts covariates exceeding
the 0.15 threshold.

\subsection{RQ2 Downstream-Contact Cohort}

RQ2 examined users located one interaction farther downstream from
the focal RQ1 users. For each matched focal user-cycle, we identified
posts and comments produced after its treatment or placebo index date.
A downstream user entered the candidate cohort upon directly replying
to one of these contributions, with the first eligible direct reply
establishing the contact timestamp.

Each downstream account contributed at most one observation per admissions
cycle. We excluded downstream users who authored or directly commented on
an anchor post during the same cycle. Eligibility required at least one
same-cycle contribution strictly before first contact and at least one
contribution strictly afterward and before June 1.

The chain-construction stage began with 8,903 matched focal user-cycles
and 209,767 eligible post-index focal-user contributions. It identified
73,413 direct-contact events and 33,191 unique downstream user-cycles.
Of these, 23,756 first contacted an exposed focal user and 9,435 first
contacted a control focal user. Requiring pre-contact and post-contact
activity yielded 19,105 user-cycles eligible for second-stage matching.

\subsection{RQ2 Propensity-Score Estimation}

For each eligible downstream user-cycle, we constructed the same 518
covariates using activity strictly before first contact. Separate
AdaBoost models within each community and admissions cycle estimated
the probability of contacting an exposed rather than a control focal
user. Each model used 200 depth-two decision trees, a learning rate
of 0.05, and ten equal-width propensity-score strata.

The RQ2 propensity models incorporated focal-user ATT weights inherited
from RQ1. A stratum was retained only when it contained at least ten
downstream users from each chain group. Treated-chain observations retained
their inherited focal-user weights. For control-chain observations,
inherited weights were multiplied by the ratio of treated-chain to
control-chain focal-weight mass within the corresponding community-cycle
propensity stratum.

The second-stage design retained 18,923 of the 19,105 eligible downstream
user-cycles across 20 propensity-score strata, comprising 13,592
treated-chain and 5,331 control-chain observations
(Table~\ref{tab:app-rq2-matching-flow}). Maximum absolute SMD decreased
from 0.2094 before matching to 0.0594 afterward.

\begin{table}[t]
\centering
\small
\setlength{\tabcolsep}{4pt}
\begin{tabular}{@{}lr@{}}
\toprule
Stage & User-cycles \\
\midrule
Downstream identified & 33,191 \\
Eligible pre-/post-contact & 19,105 \\
Retained after matching & 18,923 \\
\quad Treated-chain & 13,592 \\
\quad Control-chain & 5,331 \\
Retained strata & 20 \\
\bottomrule
\end{tabular}
\caption{RQ2 downstream cohort construction and matching.}
\label{tab:app-rq2-matching-flow}
\end{table}

\section{Outcome Operationalization}
\label{app:outcomes}

\subsection{Aggregation and Observation Windows}

All linguistic outcomes were calculated at the user-cycle-period level.
For each observation period, eligible posts and comments were ordered
chronologically and concatenated using newline separators. Each resulting
document was scored once per SVM and LIWC outcome, yielding one baseline
and one outcome score per user-cycle, irrespective of contribution count.

For RQ1, baseline documents included all observed same-cycle activity
strictly before the treatment or placebo index. Outcome documents
included activity from December 1 through May 31. For RQ2, baseline
documents included activity strictly before the downstream user's first
eligible contact, while outcome documents included activity strictly
after contact through May 31.

Word tokens were extracted as alphabetic strings permitting internal
apostrophes, while letter counts included A--Z characters irrespective
of case. Sentence units were identified by terminal periods, question
marks, or exclamation marks. Contributions without terminal punctuation
were assigned one sentence unit.

\subsection{Affective Outcomes}

Anxiety, depression, and stress were measured using previously trained support-vector-machine classifiers. Each concatenated period-level document was transformed using the corresponding stored n-gram vectorizer and passed to its classifier. We retained the native decision-function score. These scores are signed, uncalibrated margins rather than probabilities or clinical assessments. Higher values indicate greater alignment with the classifier's positive class.

\subsection{Behavioral Outcomes}

We measured five dimensions of observable participation. Let $C_u$
denote contributions, $P_u$ original posts, $R_u$ comments,
$A_u$ calendar days with activity, and $T_u$ calendar days in the
observation window. Outcomes were defined as

\begin{align}
\text{Contributions/day}_u &= \frac{C_u}{T_u},\\
\text{Posts/day}_u &= \frac{P_u}{T_u},\\
\text{Comments/day}_u &= \frac{R_u}{T_u},\\
\text{Active-day share}_u &= \frac{A_u}{T_u},\\
\text{Comment share}_u &= \frac{R_u}{C_u}.
\end{align}

For RQ1, the baseline denominator extended from the first observed
baseline contribution through the index date, while the outcome
denominator covered the full December--May observation window.
For RQ2, baseline duration extended from the first observed
contribution through contact, and outcome duration extended from
contact through May 31.

These measures capture observable contributions within the study
communities, not passive reading or browsing.

\subsection{Structural-Linguistic Outcomes}

Let $W_u$ denote word tokens, $L_u$ alphabetic characters,
$S_u$ detected sentence units, and $V_u$ unique case-normalized
word tokens in the period-level document. We defined four
structural-linguistic outcomes.

\paragraph{Readability.}
We used the Coleman--Liau grade-level index:

\begin{equation}
\operatorname{CLI}_u
=0.0588\left(100\frac{L_u}{W_u}\right)
-0.296\left(100\frac{S_u}{W_u}\right)-15.8.
\end{equation}

Higher scores indicate a higher estimated reading grade level,
not necessarily better writing.

\paragraph{Verbosity.}
Mean words per sentence was calculated as

\begin{equation}
\operatorname{Verbosity}_u = \frac{W_u}{S_u}.
\end{equation}

\paragraph{Repeatability.}
Lexical repeatability was defined as

\begin{equation}
\operatorname{Repeatability}_u
=1-\frac{V_u}{W_u}.
\end{equation}

Higher scores indicate greater lexical reuse within the
concatenated document.

\paragraph{Lexical Complexity.}
Mean letters per word was calculated as

\begin{equation}
\operatorname{Complexity}_u = \frac{L_u}{W_u}.
\end{equation}

Higher values represent longer words on average, rather than
greater cognitive complexity or functioning.

\subsection{Implications of Period-Level Concatenation}

Period-level concatenation ensured that each user-cycle contributed
one observation per outcome, regardless of contribution frequency.
However, linguistic measures may remain sensitive to text volume.
Longer documents provide more evidence to SVM vectorizers, while
repeatability is mechanically related to document length through
additional opportunities for lexical reuse.

Accordingly, SVM margins should not be interpreted as clinical
severity measures, nor repeatability as direct evidence of cognitive
functioning. Propensity-score estimation incorporated baseline
activity volume, active-day share, mean words per contribution,
and observed-history duration. Outcome models also adjusted for
the corresponding pre-index outcome. These adjustments account
for observed baseline differences but do not eliminate sensitivity
to subsequent contribution volume.

\section{RQ1 Exposure-Effect Heterogeneity Across Admissions Cycles}
\label{app:rq1-cycle-heterogeneity}

\begin{table}[ht]
\centering
\small
\setlength{\tabcolsep}{4pt}
\begin{tabular}{lrrr}
\toprule
Outcome & Wald $\chi^2(2)$ & $p$ & $q_{BH}$ \\
\midrule
\multicolumn{4}{l}{\textit{Affective outcomes}} \\
Anxiety SVM & $1.137$ & $0.566$ & $0.955$ \\
Depression SVM & $0.257$ & $0.879$ & $0.955$ \\
Stress SVM & $1.397$ & $0.497$ & $0.955$ \\
LIWC positive emotion & $7.753$ & $0.021$ & $0.166$ \\
LIWC negative emotion & $1.308$ & $0.520$ & $0.955$ \\
LIWC anxiety & $0.269$ & $0.874$ & $0.955$ \\
LIWC anger & $0.092$ & $0.955$ & $0.955$ \\
LIWC sadness & $0.869$ & $0.648$ & $0.955$ \\

\addlinespace
\multicolumn{4}{l}{\textit{Behavioral outcomes}} \\
Contributions/day & $4.840$ & $0.089$ & $0.152$ \\
Posts/day & $1.424$ & $0.491$ & $0.613$ \\
Comments/day & $4.784$ & $0.091$ & $0.152$ \\
Active-day share & $7.882$ & $0.019$ & $0.097$ \\
Comment share & $0.248$ & $0.884$ & $0.884$ \\

\addlinespace
\multicolumn{4}{l}{\textit{Structural-linguistic outcomes}} \\
Readability & $0.962$ & $0.618$ & $0.799$ \\
Verbosity & $2.085$ & $0.352$ & $0.705$ \\
Repeatability & $2.254$ & $0.324$ & $0.705$ \\
Complexity & $0.450$ & $0.799$ & $0.799$ \\
\bottomrule
\end{tabular}
\caption{\textbf{RQ1 cycle heterogeneity in r/MBA.}
Omnibus Wald tests of equality of baseline-adjusted exposure
coefficients across the three admissions cycles. Benjamini--Hochberg
correction was applied within outcome family. Two outcomes were
nominally significant at $p<.05$, but none remained significant
at $q_{BH}<.05$.}
\label{tab:app-rq1-cycle-mba}
\end{table}


\begin{table}
\centering
\small
\setlength{\tabcolsep}{4pt}
\begin{tabular}{lrrr}
\toprule
Outcome & Wald $\chi^2(2)$ & $p$ & $q_{BH}$ \\
\midrule
\multicolumn{4}{l}{\textit{Affective outcomes}} \\
Anxiety SVM & $1.544$ & $0.462$ & $0.743$ \\
Depression SVM & $1.975$ & $0.372$ & $0.743$ \\
Stress SVM & $1.181$ & $0.554$ & $0.743$ \\
LIWC positive emotion & $0.689$ & $0.708$ & $0.743$ \\
LIWC negative emotion & $0.595$ & $0.743$ & $0.743$ \\
LIWC anxiety & $2.035$ & $0.361$ & $0.743$ \\
LIWC anger & $0.813$ & $0.666$ & $0.743$ \\
LIWC sadness & $1.744$ & $0.418$ & $0.743$ \\

\addlinespace
\multicolumn{4}{l}{\textit{Behavioral outcomes}} \\
Contributions/day & $1.466$ & $0.480$ & $0.984$ \\
Posts/day & $0.633$ & $0.729$ & $0.984$ \\
Comments/day & $1.681$ & $0.432$ & $0.984$ \\
Active-day share & $0.047$ & $0.977$ & $0.984$ \\
Comment share & $0.032$ & $0.984$ & $0.984$ \\

\addlinespace
\multicolumn{4}{l}{\textit{Structural-linguistic outcomes}} \\
Readability & $0.555$ & $0.758$ & $0.820$ \\
Verbosity & $4.582$ & $0.101$ & $0.405$ \\
Repeatability & $1.218$ & $0.544$ & $0.820$ \\
Complexity & $0.397$ & $0.820$ & $0.820$ \\
\bottomrule
\end{tabular}
\caption{\textbf{RQ1 cycle heterogeneity in r/gradadmissions.}
Omnibus Wald tests of equality of baseline-adjusted exposure
coefficients across the three admissions cycles. Benjamini--Hochberg
correction was applied within outcome family. No test was
significant at $q_{BH}<.05$.}
\label{tab:app-rq1-cycle-gradadmissions}
\end{table}

\appendix

\end{document}